\pdfoutput=1

\documentclass[11pt]{article}

\usepackage{acl}

\usepackage{times}
\usepackage{latexsym}

\usepackage{soul}
\usepackage{url}
\usepackage{hyperref}
\usepackage{caption}
\usepackage{graphicx}
\usepackage{amsmath}
\usepackage{booktabs}
\usepackage{makecell}
\usepackage{subfigure}
\usepackage{threeparttable}
\usepackage{multirow}
\usepackage{multicol}
\usepackage{xcolor}
\usepackage[normalem]{ulem}
\usepackage[switch]{lineno}
\usepackage{bbm}
\usepackage{amssymb}
\usepackage{enumitem}

\usepackage{lipsum}

\usepackage[T1]{fontenc}

\usepackage[utf8]{inputenc}

\usepackage{microtype}

\newcommand{\dataset}{\textsc{DialMed}}
\newcommand{\model}{DDN}

\title{{\dataset}: A Dataset for Dialogue-based Medication Recommendation}

\author{
    Zhenfeng~He$^{1}$\footnotemark[1],
    Yuqiang~Han$^{1}$\footnotemark[1], 
    Zhenqiu~Ouyang$^{3}$, 
    Wei~Gao$^{4}$, 
    Hongxu~Chen$^5$, \\
    {\bf Guandong~Xu$^5$}, 
    {\bf Jian~Wu$^{2}$\footnotemark[2]} \\
    $^1$College of Computer Science and Technology, Zhejiang University\\
    $^2$School of Public Health, Zhejiang University    $^3$Polytechnic Institute, Zhejiang University\\
    $^4$Ningbo Institute of Technology, Zhejiang University\\
    $^5$University of Technology Sydney\\
    \{hezf, hyq2015, oyzq, gw, wujian2000\}@zju.edu.cn \\
    \{hongxu.chen, guandong.xu\}@uts.edu.au
}

\begin{document}

\maketitle

\renewcommand{\thefootnote}{\fnsymbol{footnote}}
\footnotetext[1]{Both authors contributed equally to this research.}
\footnotetext[2]{Corresponding author.}
\renewcommand{\thefootnote}{\arabic{footnote}}

\begin{abstract}

Medication recommendation is a crucial task for intelligent healthcare systems. Previous studies mainly recommend medications with electronic health records~(EHRs).
However, some details of interactions between doctors and patients may be ignored or omitted in EHRs, which are essential for automatic medication recommendation.
Therefore, we make the first attempt to recommend medications with the conversations between doctors and patients.
In this work, we construct {\dataset}, the first high-quality dataset for medical dialogue-based medication recommendation task.
It contains $11,996$ medical dialogues related to $16$ common diseases from $3$ departments and $70$ corresponding common medications.
Furthermore, we propose a \textbf{D}ialogue structure and \textbf{D}isease knowledge aware  \textbf{N}etwork~(\textbf{{\model}}), where a QA Dialogue Graph mechanism is designed to model the dialogue structure and the knowledge graph is used to introduce external disease knowledge. 
The extensive experimental results demonstrate that the proposed method is a promising solution to recommend medications with medical dialogues.
The dataset and code are available at \url{https://github.com/f-window/DialMed}.

\end{abstract}

\section{Introduction}
The outbreak of COVID-19 has challenged the healthcare systems and led to millions of patients facing delays in diagnosis and treatment.
As an essential complement to the traditional face-to-face medicine, telemedicine relieved the therapeutic stress caused by the diversion of medical resources.
According to the report of WeDoctor\footnote{\url{https://www.guahao.com/}}, an online health consultation platform in China, about $1.2$ million patients conducted online medical consultations during the COVID-19 Pandemic.
Telemedicine can increase the availability of medical treatment, reduce healthcare costs, and improve the quality of care. Consequently, it has attracted increasing attention due to its vast application potential.
\begin{figure}[!tbp]
    \centering
    \includegraphics[scale=0.4]{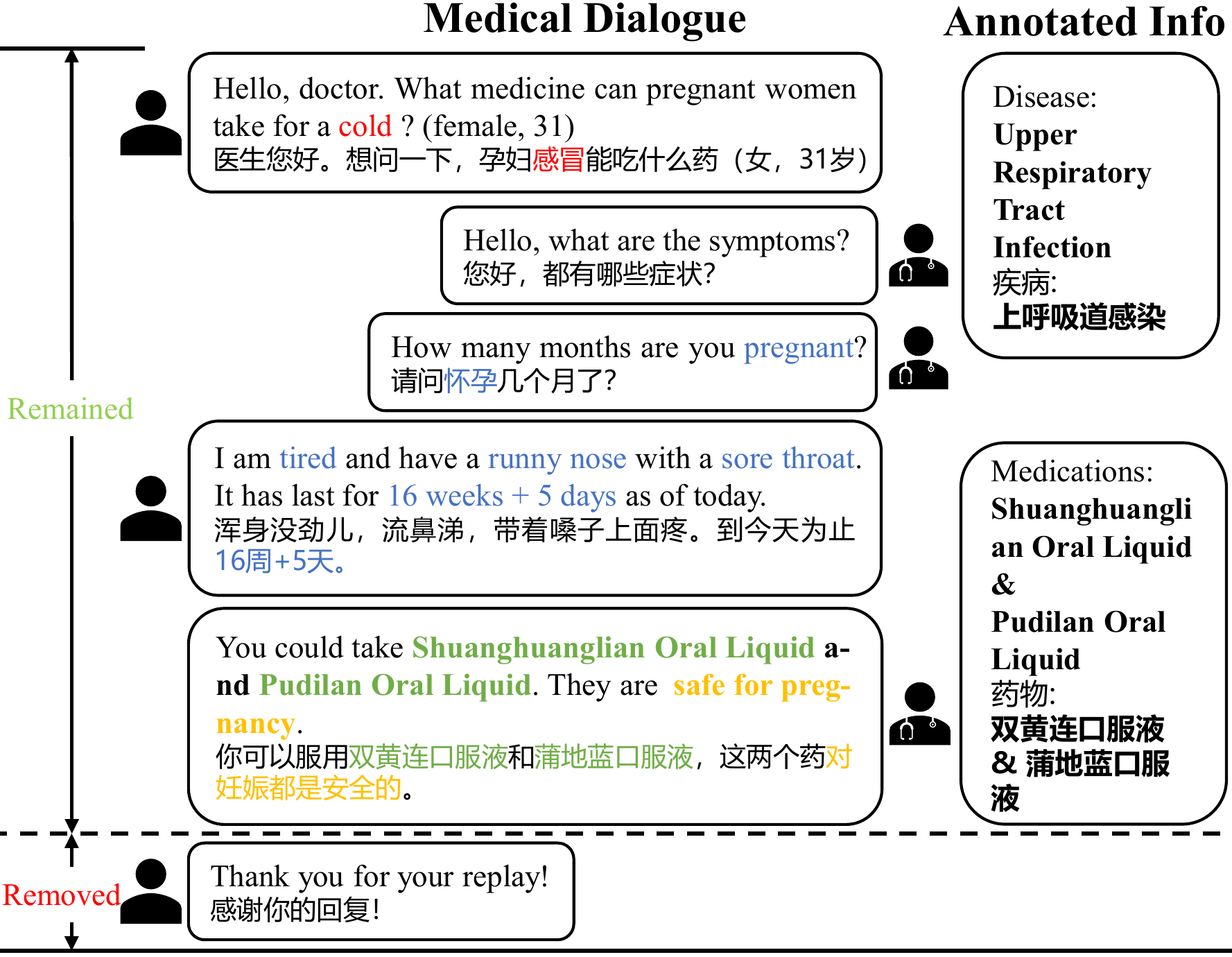}
    \caption{A typical medication consultation dialogue. Here, the disease is \textbf{\textit{Upper~Respiratory~Tract~Infection}}, and the medication is \textbf{\textit{Shuanghuanglian~Oral~Liquid}} and \textbf{\textit{Pudilan~Oral~Liquid}}.
    }
    \label{fig:example}
\end{figure}%

Our study found that around $31\%$ of online consultations are about what medications the patients should take based on their current conditions\footnote{Refer to Appendix \ref{sec:ratio_medication} for details of statistic.}.
Figure~\ref{fig:example} demonstrates a typical medication consultation dialogue. The patient reported the health issues initially, with some personal information, such as gender and age. Then the doctor asked for further information (\textit{e.g.}, symptoms and disease history) about the patient. Finally, the doctor provided medication advice based on the gathered information and clinical experience. 

Existing studies on medication recommendation are primarily based on  EHRs~\cite{zhang2017leap,shang2019gamenet,an2021prediction}, accumulatively collected according to a diagnostic procedure in clinics. 
However, the doctors will omit some details of interactions with patients in EHRs, which are essential for the automatic medication recommendation. 
Compared to EHRs, medical dialogues retain original interactions between doctors and patients, containing more rich information.
To this end, medical dialogue-based medication recommendation is a promising and challenging task.

Therefore, in this work, we study the new task, namely dialogue-based medication recommendation. 
Due to the lack of available datasets, we firstly construct a high-quality online medical dialogues dataset~({\dataset}) for this task. It contains $11,996$ consultation dialogues, $16$ diseases from $3$ different departments, and $70$ related common medications. 

Then, to further advance the research of this task, we propose a \textbf{D}ialogue structure and \textbf{D}isease knowledge aware \textbf{N}etwork~(\textbf{{\model}}).
In {\model}, for the input dialogue, we first utilize a pre-trained language model to extract the semantic information of each utterance. A mechanism named QA Dialogue Graph is designed to understand the questions\&answers implied in utterances, and then we apply graph attention network on this QA graph to get the dialogue embedding.
Meanwhile, for the input disease, we use its identity to query the entity in a knowledge graph CMeKG\footnote{\url{http://cmekg.pcl.ac.cn/}}, and input the dialogue embedding to a graph attention network to get contextual disease embedding.
The two embeddings are fused to make the medication prediction.
Moreover, we conduct extensive experiments to show that the proposed method can effectively recommend medications with medical dialogues.

Our contributions can be summarized as follows:
\begin{itemize}[itemsep=0in,topsep=0in,leftmargin=0.3in]
    \item We construct the first high-quality human-annotated dialogue dataset for dialogue-based medication recommendation task.
    \item  We propose a novel medication recommendation framework which models dialogue structure with QA Dialogue Graph and introduces external disease knowledge.
    \item  We conduct extensive experiments to demonstrate {\model} can extract the essential information to make medication recommendation effectively.
\end{itemize}

\section{Related Work}
\label{sec:related}

\paragraph{Medication Recommendation.}
Existing medication recommendations are mainly based on EHRs. It could be categorized into instance-based and longitudinal-based recommendation methods~\cite{shang2019gamenet}.
Instance-based methods are based on the current health conditions extracted from recent visit~\cite{zhang2017leap,wang2019order}. For example, \cite{zhang2017leap} proposed a multi-instance multi-label learning framework to predict medication combination based on patient's current diagnoses. Longitudinal-based methods leverage the temporal dependencies among clinical events~\cite{choi2016retain,le2018dual,shang2019gamenet,shang2019pre,wang2020seqmed,he2020attention,wang2021adversarially,yang2021safedrug}. Among them, \cite{shang2019pre} combined the power of graph neural networks and BERT for medication recommendation. \cite{yang2021safedrug} proposed a drug-drug interactions (DDI)-controllable drug recommendation model to leverage drugs' molecule structures and model DDIs explicitly.

Unlike the work mentioned above, dialogue-based medication recommendation task is more challenging in practice due to the noisy and sparse data.
Because of the privacy issue, it is difficult to get historical dialogues of a patient on online consultation platforms.
So we perform the medication recommendation solely based on the current medical dialogues.

\paragraph{Graph Neural Networks.}
Graph neural networks have attracted a lot of attention for processing data with graph structures in various domains~\cite{zhou2020graph}. For example, \cite{kipf2016semi} proposed the graph convolutional networks (GCN). With integration of attention mechanisms, graph attention networks(GAT) \cite{velivckovic2017graph} has become one of the most popular methods in graph neural networks.

Recently, some works have applied GAT to the dialogue modeling. \cite{chen2020schema} used Graph attention and recurrent GAT to fully encode dialogue utterances, schema graphs, and previous dialogue states for dialogue state tracking. \cite{qin2020co} proposed a co-interactive GAT layer to simultaneously solve both dialog act recognition and sentiment classification task.
In this work, we utilize GAT to model the intra- and inter-speaker correlations to propagate semantic on the QA Dialogue Graph and extend the GAT on knowledge graph to introduce external knowledge.
\section{Corpus Description}
\label{sec:corpus}
In this section, we introduce the construction details and statistics of {\dataset}, and its comparison with other studies.

\subsection{Construction Details}
Our dataset is collected from Chunyu-Doctor\footnote{\url{https://www.chunyuyisheng.com/}}, which is a popular Chinese medical consultation website for doctors and patients. 
The conversations between doctors and patients contain rich but complex information, mainly related to the patients' current conditions.
The diagnosed diseases and symptoms both are indispensable for accurate medication recommendation.
Considering the complexity of the symptoms, we decide to utilize information from \textit{explicit disease} and \textit{implicit symptoms} in this paper. 
So we annotate the diagnosed diseases and recommended medications~(replaced with a mask token to keep the original dialogue structure).
For the example in Figure~\ref{fig:example}, we annotate the disease \textit{Upper Respiratory Tract Infection}, and replace the medications \textit{Shuanghuanglian~Oral~Liquid} and \textit{Pudilan~Oral~Liquid} with special token \textit{[MASK]}. 
Moreover, the future utterances after the point of recommendation are removed to make {\dataset} more realistic, as the decision of doctors should not be influenced by future contexts.

The procedure of annotation consists of two parts, labeling and normalization of medications and diseases.
First, we select $16$ common diseases and the corresponding common medications from $3$ departments (i.e., respiratory, gastroenterology, and dermatology) with the guidance of a doctor.
These diseases have abundant medication consultations online. 
Then three annotators with relevant medical backgrounds are involved. 
Each dialogue is annotated by two annotators and will be further judged by another one if there is any inconsistency.
The annotation consistency, i.e., the Cohen's kappa coefﬁcient~\cite{fleiss1973equivalence} of the labelled dialogues is $88.4\%$. For the quality of dataset, conversations containing unsuitable medications for patients would be discarded.

Secondly, we normalize the medications since there are many generic names, trade names, or colloquial expressions for the same drug in dialogues. Specifically, different brands of the same drug are grouped into one cluster and normalized as a common name from DXY Drugs Database\footnote{\url{http://drugs.dxy.cn/}}. For example, \textbf{Omeprazole enteric-coated tablet} and \textbf{Omeprazole tablet} are normalized to \textbf{Omeprazole}. Similarly, we normalize the different names of diseases into ICD-10 standard names. 
The dialogues, hard to give diagnosed diseases or given diseases out of our scope, would be marked as a special placeholder, \textbf{None or Others}.

\subsection{Dataset Statistics}
\begin{table*}[!tbp]
     \setlength{\belowcaptionskip}{0.3cm}
     \small
     \resizebox{0.99\linewidth}{!}{
     \centering
     \begin{tabular}{l|cccccc}    
     \toprule    
       Dataset & \#Task & \#Domain & \#Disease & \#Dialogue & \#Avg. Turn & \#Annotation\\    
     \midrule   
     MZ\cite{wei2018task} & Diagnosis & Pediatrics & 4 & 710 & - & Man.\\   
     DX\cite{xu2019end}  &  Diagnosis & Pediatrics & 5 & 527 & 5.34 & Man.\\  
     CMDD\cite{lin2019enhancing} & Diagnosis & Pediatrics & 4 & 2,067 & 42.09 & Man.\\  
     
     \midrule
     
     SAT\cite{du2019extracting} & Extraction & 14 & - & 2,950 & - & Man.\\ 
     MIE\cite{zhang2020mie} &  Extraction & Cardiology & 6 & 1,120 & 16.19 & Man.\\
     MSL\cite{shi2020understanding} & Extraction & Pediatrics & 5 & 2,652 & - & Man.\\
     MedDG\cite{liu2020meddg} & Extraction & Gastroenterology & 12 & 17,864 & 21.60 & Man.\& Semi-Auto.\\
     
     \midrule
     
     COVID-EN\cite{yang2020generation} & Generation & COVID-19 & 1 & 603 & 8.7 & None\\
     COVID-CN\cite{yang2020generation} & Generation & COVID-19 & 1 & 1088 & 2.0 & None\\
     MedDialog-EN\cite{zeng2020meddialog} & Generation & 51 & 96 & 257,332 & 2 & None\\
     MedDialog-CN\cite{zeng2020meddialog} & Generation & 29 & 172 & 3,407,494 & 3.3 & None\\
     Chunyu\cite{lin2021graph} & Generation & - & 15 & 12,842 & 24.7 & Rule\\
     KaMed\cite{10.1145/3404835.3462921} & Generation & 100 & - & 63,754 & 11.62 & None\\
     ReMeDi\cite{yan2022remedi} & Diag.\&Ext.\&Gene. & 30 & 491 & 1,557 & 16.34 & Man.\\
     \midrule
     {\dataset}(ours) & Medication & R\&G\&D & 16 & 11,996 & 10.94 & Man.\\
     \bottomrule   
     \end{tabular}} 
     \caption{Comparison between our dataset and other related medical dialogue datasets. Extraction, Generation and Medication mean information extraction, dialogue generation and medication recommendation separately. R\&G\&D, Man. and Semi-Auto are abbreviations of  Respiratory\&Gastroenterology\&Dermatology, Manual and Semi-Automated respectively.}
    \label{tab:dataset_comparison}
\end{table*}

\begin{table}[!tp]
    \setlength{\tabcolsep}{1.4pt}
    \renewcommand\arraystretch{1.1}
    \small
    \centering
    \resizebox{0.99\linewidth}{!}{
        \begin{tabular}{lcccccccc}    
        \toprule    
          & \#Dial. & \#Dise. & \#Med. & Avg.M & Avg.T & Max.T & Avg.U & Max.U\\
          \midrule 
         Resp. & 4,859 & 4 & 45 & 2.06 & 10.76 & 52 & 18.18 & 374  \\
         Gastro. & 3,818 & 9 & 39 & 1.88 & 13.05 & 58 & 16.70 & 463  \\
         Derma. & 3,319 & 3 & 27 & 1.62 & 8.77 & 44 &  18.82 & 453   \\
         \hline 
         Total & 11,996 & 16 & 70 & 1.88 & 10.94 & 58 & 17.76 & 463  \\
        \hline \hline 
         Train. & 9,605 & 16 & 70 & 1.88 & 10.95 & 58 & 17.74 & 463  \\
         Dev. & 1,192 & 16 & 70 & 1.89 & 11.25 & 49 & 17.45 & 298  \\
         Test. & 1,199 & 16 & 70 & 1.89 & 10.58 & 42 &  18.27 & 293   \\
         \bottomrule
        \end{tabular} }
     \caption{Data statistics of {\dataset}. M, T, and U represent medicine, dialogue turns, and utterance.}  
    \label{tab:dataset}
\end{table}%

\begin{figure}[!tbp]
    \centering
    \subfigure[The frequency of medications.]{
        \includegraphics[scale=0.5]{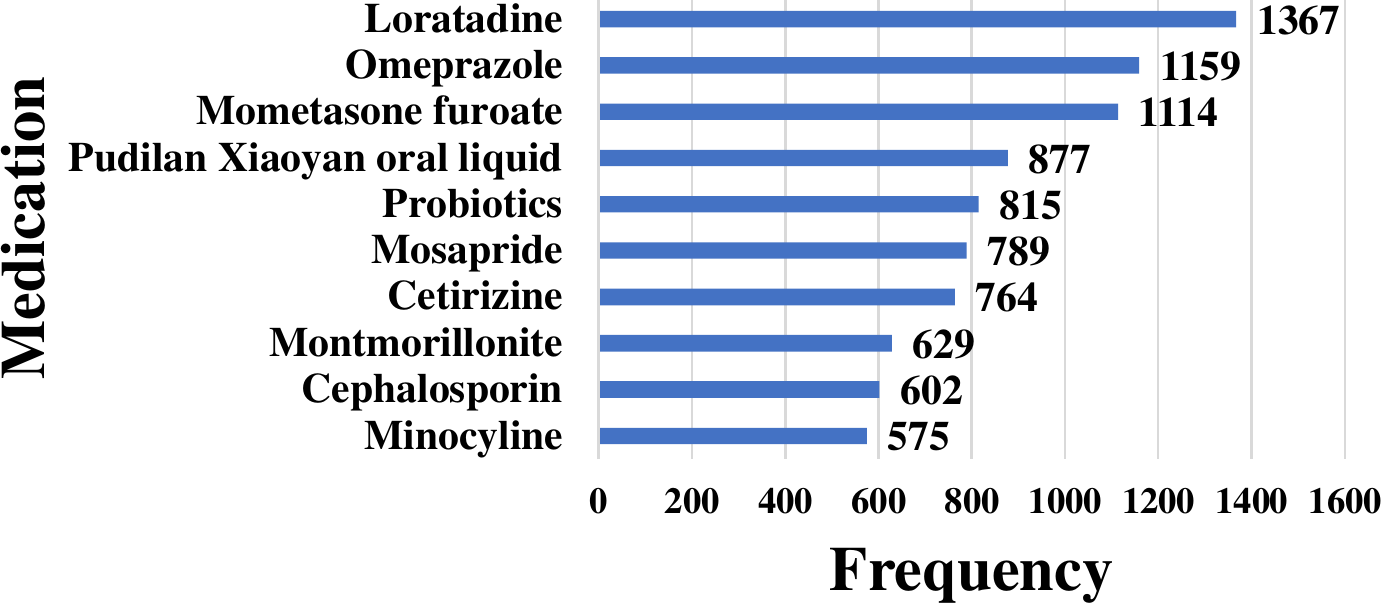}
        \label{fig:medication}
    }
    \quad    
    \subfigure[The frequency of diseases.]{
	\includegraphics[scale=0.5]{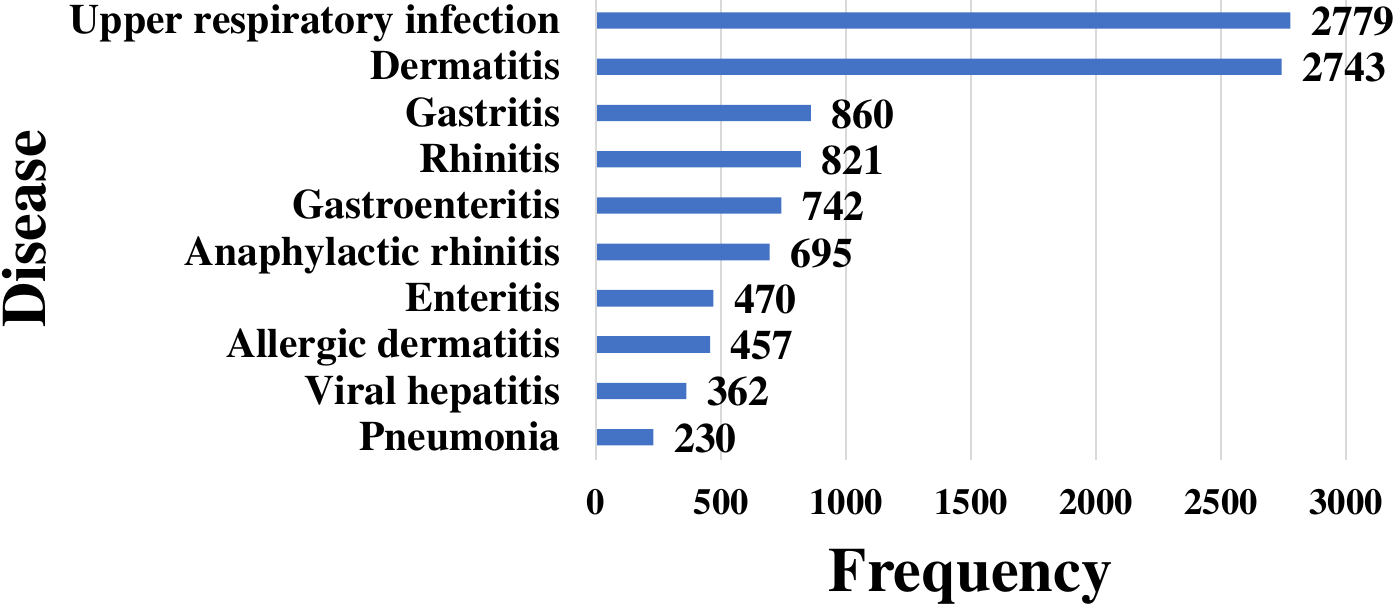}
	\label{fig:disease}
    }
    \caption{The frequencies of medications and diseases. Top 10 are exhibited for the constraint of space.}
\end{figure}

Top of Table~\ref{tab:dataset} summarizes the statistics of {\dataset}. The scenario of dialogues in the dataset is similar to outpatient procedure, so the number of medicines per dialogue is relatively small. 
Then, the frequency of medications and diseases are shown in Figure~\ref{fig:medication} and Figure~\ref{fig:disease} respectively. The distributions of quantity demonstrate that {\dataset} aligns with the real-world case.

Compared to the other medical dialogue datasets in Table~\ref{tab:dataset_comparison}, our dataset has three advantages: (1) {\dataset} has the largest volume among the manual annotation datasets, as unlabeled datasets are mainly constructed for the task of dialogue generation. (2) Though the future contexts after recommendation are removed, the average number of dialogue turns in {\dataset} still remains high compared to other datasets. It is mainly benefited from our evaluation for inclusion of short dialogues in {\dataset} during the labeling process. (3) We carefully choose the fields suitable for medication recommendation and avoid coarsely expanding the scope of medical domains, which makes {\dataset} have a higher quality. 

The panoramas of medications \& diseases' frequency could be found in Appendix \ref{sec:complete_corpus_statis}.

\subsection{ The comparison with other studies  }
To our best knowledge, {\dataset} is the first dataset for the medication recommendation based on medical dialogues. It has the following differences with the existing work.

\paragraph{Dataset} 
Medical dialogue has attracted increasing attention in recent years.
Although there are medication mentions in many medical dialogue datasets, the distributions are fragmentary and the authors do not categorize and normalize these drug mentions which would lead to label explosion. For instance, medication mentions, \textbf{Omeprazole entericcoated tablet}, \textbf{Omeprazole tablet} and \textbf{Omeprazole}, which may occur in dialogues would be three classes without normalization. In fact, they are essentially equivalent in the eyes of doctors. By contrast, we reduce the complexity caused by the doctors' preferences for different brands through categorization and normalization. {\dataset} is developed for drug recommendation.

\paragraph{Task}
Drug recommendation is a sub-task of medical diagnosis. According to patients' questions, the objectives of current diagnosis systems are to generate the optimal clinical responses which may be intended as one of greeting, inquiry or diagnosis. Even if there contains drug mentions in responses, it is just one of the system's options. Drug recommendation is a key task and requires specialized dataset. {\dataset} goes a step forward.

\paragraph{Scenario}
There are remarkable distinctions between {\dataset} and MIMIC-III~\cite{johnson2016mimic}, an EHR database which is relied on in current medication recommendation study. The scenario of the former is outpatient procedure while the data from the latter is generated from Intensive Care Units (\textbf{ICU}). In MIMIC-III, for example, the number of medications is $145$, the average number of medications in each visit is $8.80$,  and the average number of diagnosis in each visit is $10.51$. In contrast, the labels in medical dialogues are relatively sparse, leading to a more challenging task.
\section{Our Approach}
\label{sec:model}
In this section, we first introduce the dialogue-based medication recommendation task, and then describe the proposed {\model} in detail.

\subsection{Problem Formulation}
\label{subsec:problem}
In the online medical dialogue setting, each dialogue consists of a sequence of utterances from the patient and the doctor. Formally, each dialogue can be represented as \(\mathcal{D}_n=\{u_1,u_2,...,u_{|\mathcal{D}_n|}\}\), where \(n \in \{ 1,2,...,N \} \), \(N\) denotes the total number of dialogues in the dataset, and \(|\mathcal{D}_n|\) represents the number of utterances in a dialogue \(\mathcal{D}_n\). 
Each utterance can be represented as \( u_i=\{w_i^1,...,w_i^j,...,w_i^{|u_i|}\}\), where \(w_i^j\) is the $j$-th word in $u_i$ and \(|u_i|\) denotes the number of words in \(u_i\). 
We collect all the diseases and medications mentioned in the dataset to construct a disease corpus \(\mathcal{S}\) and medication corpus \(\mathcal{M}\). 
To avoid notation clutter, we hereinafter remove the subscript \(n\) as we only consider a single dialogue instance.
Formally, given the consultation dialogue $\mathcal{D}$ and the diagnosed disease $d$ as inputs, dialogue-based medication recommendation aims to recommend potential treatment medications $\mathbf{y}$ in $\mathcal{M}$, where $\mathbf{y}  \in \{0,1\}^{|\mathcal{M}|}$.

\subsection{Model Overview}
\begin{figure*}
    \centering
    \includegraphics[width=0.9\textwidth]{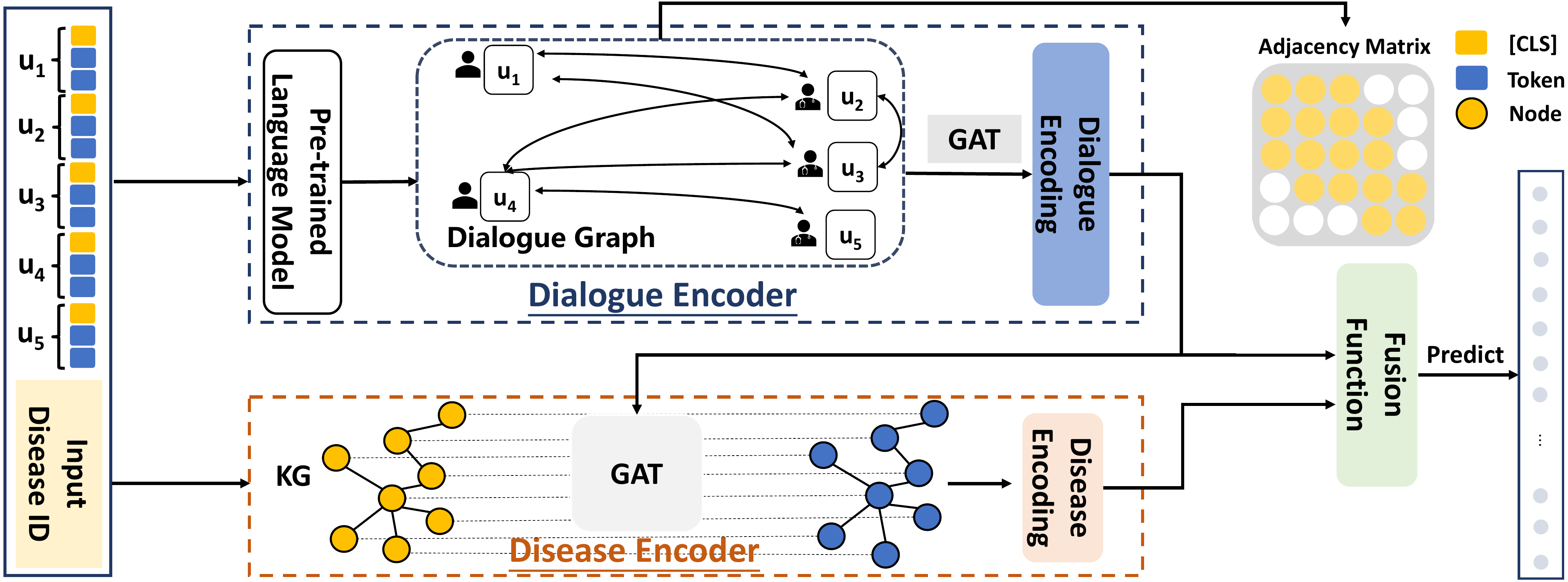}
    \caption{The framework of the proposed {\model} for dialogue-based medication recommendation.}
    \label{fig:model}
\end{figure*}%
The proposed end-to-end framework is presented in Figure~\ref{fig:model}, consisting of two parts: 
(1) Dialogue Encoder, encoding the medical dialogues between patient and doctor by comprehensively capturing the semantic information and dialogue structure. 
(2) Disease Encoder, incorporating external medical knowledge based on the disease information from the dialogue and knowledge graph.

\subsection{Dialogue Encoder}
Dialogues contain two types of important information: (1) the rich semantic information, (2) strong structural correlations between utterances.

\paragraph{Utterance Encoding}
Pre-trained language models~(\textit{e.g.}, RoBERTa) are utilized to capture the semantic information in utterances.
First, special tokens [CLS] (capturing utterance representation) and [SEP] (separating different utterances) are inserted at the beginning and end of each utterance token sequence $u_i$. 
Then the position embedding of each token in a utterance is calculated.
In addition, two types of speaker embeddings (\textit{i.e.}, \textit{Doctor} and \textit{Patient}) are proposed to make model aware of the speaker role of the utterance.
The model takes the sum of three embeddings as input and outputs the representation of [CLS] as the utterance embedding $\mathbf{h}$.
So a dialogue $\mathcal{D}$ can be represented as $\mathbf{h}_D =\{\mathbf{h} _1, \mathbf{h}_2, ..., \mathbf{h}_{|\mathcal{D}|}\}$.

\paragraph{QA Dialogue Graph}
In medical conversations, the interactions between doctors and patients tend to be in the form of questions and answers. For example, in Figure~\ref{fig:model}, the doctor asked two questions in \(u_2\) and \(u_3\), and the patient gave the answers in \(u_4\). So it's important to capture the structure of QA pairs in conversation in order to understand the whole medical dialogue. 
We propose a new method to model the dialogue based on the observation that there is a high possibility of question-and-answer relations between adjacent utterances.

Specifically, we design a mechanism named QA Dialogue Graph, where each utterance is represented as a node in graph, and consecutive utterances spoken by the same speaker is represented as a block, e.g., \(u_2\) and \(u_3\) constitute a block with two nodes, and \(u_4\) is another block with one node.
Then the constructions of edges between nodes can be defined as follows: 
\begin{itemize}
\item Within a block, each node connects with all other nodes in the block. This represents the intra-speaker correlation and ensures the information from the same speaker propagates among utterances within a local context.
\item For two adjacent blocks, each node in a block connects with all nodes in the other block. This represents the inter-speaker correlation and ensures the information flow between doctors and patients within consecutive contexts.
\end{itemize}

An example of the adjacency matrix of the dialogue is shown in Figure~\ref{fig:model}. In general, when compared to previous works on dialogue modeling, QA Dialogue Graph has two advantages. Firstly, the construction of graphs does not require additional supervised information~\cite{joshi2021dialograph,feng2020dialogue}. Secondly, our method comprehensively captures the structural and semantic information of QA pairs, which is key to understanding conversations~\cite{qin2020co,shen2021directed}.

\paragraph{Dialogue Encoding}
GAT is employed to automatically aggregate semantic and structure features on QA Dialogue Graph.
In particular, the $l$-th layer representation of a vertex can be computed as: 
\begin{equation} \small
	\mathbf{h} _i^{(l)} = \sigma (\sum_{j \in \mathcal{N} _i} \alpha_{ij} W_h \mathbf{h} _j^{(l-1)} )
\end{equation}
where \(\mathcal{N} _i\) is the ﬁrst-order neighbors of vertex \(i\), \(W_h \in \mathbbm{R}^{d_l \times d_{l-1}}\) is a trainable weight matrix, and \(\sigma\) is a nonlinear activation function. 
The weight \(\alpha_{ij}\) which determines the relatedness between two vertices can be calculated following~\cite{velivckovic2017graph}:
\begin{equation} \small
\alpha _{ij}=\frac{\mathrm{exp} \left ( \sigma (\mathbf{a}^TW_h[\mathbf{h} _i\left |  \right | \mathbf{h} _j]) \right )}
 { {\textstyle \sum_{k \in \mathcal{N} _i}^{}} \mathrm{exp} \left ( \sigma (\mathbf{a}^TW_h[\mathbf{h} _i\left |  \right | \mathbf{h} _k]) \right )}
\end{equation}
where \(\mathbf{a} \in \mathbbm{R}^{2d_l}\) is a trainable weight matrix, and $\sigma$ is the $\mathrm{LeakyReLU}$ activation function.
Finally, we apply the attention pooling on nodes embedding to obtain the dialogue representation $\mathbf{h}_\mathcal{D}$, where $\mathbf{W}_a$ is a learnable parameter and $\mathbf{h} ^{(l)}$ is the representation of utterances after $l^{th}$ layer.
\begin{equation} \small
    \widehat{\alpha} = softmax(\mathbf{W}_a \mathbf{h}^{(l)}) 
\end{equation}
\begin{equation} \small
    \mathbf{h}_\mathcal{D} = \sum_{i} \widehat{\alpha} _i \mathbf{h} _i^{(l)}
\end{equation}%

\subsection{Disease Encoder}
Disease knowledge is crucial for delivering accurate medication recommendation.
In this paper, we incorporate knowledge from CMeKG, a high-quality Chinese medical knowledge graph. TransR~\cite{wang2019kgat} is utilized to get the initial entities embedding.
Given a disease \(d\), we first identify the corresponding entity in CMeKG, and then a KG subset with $K$ hops starting from the disease entity is sampled randomly, finally the GAT network is used to get the disease embedding under the dialogue context.
 
Here, we fuse the entity, relation and dialogue information to get the attention scores: 
\begin{equation} \small
\beta _{ij}=\frac{\mathrm{exp} \left ( \sigma (\mathbf{a}^T[W[\mathbf{h}_i,\mathbf{h}_j] \left | \right |  W_r\mathbf{r}_{\varphi} \left | \right |  W_D\mathbf{h}_\mathcal{D}]) \right )}
 {{\textstyle \sum_{j \in \mathcal{N} _i}^{}} \mathrm{exp} \left ( \sigma (\mathbf{a}^T[W[\mathbf{h}_i,\mathbf{h}_j] \left | \right |  W_r\mathbf{r}_{\varphi} \left | \right |  W_D\mathbf{h}_\mathcal{D}]) \right )}
\end{equation}
where $\sigma$ is the $\mathrm{LeakyReLU}$ function, $\mathbf{h}_i$, $\mathbf{h}_j$ and $\mathbf{r}_\varphi$ are the embeddings of node $i$, $j$ and their relation separately. And $W$, $W_r$, and $W_D$ are learnable weights to transform node, relation and dialogue embeddings, respectively. 
Then the $l$-th layer of disease embedding can be obtained as follows: 
\begin{equation} \small
	\mathbf{s} _i^{(l)} = \sigma (\sum_{j \in \mathcal{N} _i} \beta_{ij} W_k \mathbf{h} _j^{(l-1)} )
\end{equation}
The contextual embedding of last layer is the disease $d$'s representation, denoted by~$\mathbf{s}_d$. 

For dialogues with None or Others placeholder rather than a disease label, a learnable vector~$\hat{\mathbf{s}}_d$ would be assigned to~$\mathbf{s}_d$.

\subsection{Model Inference and Optimization}
The dialogue \(\mathbf{h}_\mathcal{D}\) and disease \(\mathbf{s}_d\) are fused by the \textit{fusion function} to make prediction.
In this work, we concatenate them and then fed it into decoder to make the medication prediction as follows: 
\begin{equation} \small
\mathbf{y}  = \sigma (W_o[\mathbf{h}_\mathcal{D};\mathbf{s}_d]+\mathbf{b}_o)
\end{equation}
where $W_o \in \mathbbm{R}^{|\mathcal{M}| \times 2d}$ and  $\mathbf{b}_o \in \mathbbm{R}^{|\mathcal{M}|}$ are trainable weight matrices for the decoder, $ \sigma$ is the \textrm{sigmoid} activation function.
Here, we reserve all the candidates whose probability is higher than the threshold of $0.5$ as the recommended treatment medication combination.

Since medication combination recommendation is treated as a multi-label classification task~\cite{shang2019gamenet,yang2021safedrug}, we utilize the binary cross-entropy loss as the objective function, which can be formulated as:
\begin{equation} \small
	\mathcal{L}  = - \sum_{i=1}^{|\mathcal{D}|}\sum_{j=1}^{|\mathcal{M}|} (y_j^{(i)}\log\widehat{y}_j^{(i)}  + (1-y_j^{(i)})\log(1-\widehat{y}_j^{(i)}))
\end{equation}%
where \(|\mathcal{D}|\) is the number of dialogues in the training set, \(|\mathcal{M}|\) is the number of medications.
\(y_j^{(i)}\) is the ground truth label which equals $1$ if medication \(j\) is prescribed by the doctor in dialogue \(i\), and 0 otherwise. \(\widehat{y}_j^{(i)}\) is the predicted probability of recommending medication \(j\). 
\section{Experiments}
\label{sec:experiments}
\begin{table*}[!tbp]
    \renewcommand\arraystretch{1.3}
    \resizebox{0.99\linewidth}{!}{
    \centering
    {\begin{tabular}{llcccccccc}
        \toprule
          &  & \multicolumn{2}{c}{\textbf{All Data}} & \multicolumn{2}{c}{\textbf{Respiratory}} & \multicolumn{2}{c}{\textbf{Gastroenterology}} & \multicolumn{2}{c}{\textbf{Dermatology}} \\ \hline
        Type of Model & Model & \multicolumn{1}{c}{Jaccard} & F1 & \multicolumn{1}{c}{Jaccard} & F1 & \multicolumn{1}{c}{Jaccard} & F1 & \multicolumn{1}{c}{Jaccard} & F1 \\ \hline
        Statistics & TF-IDF\cite{salton1988term} & \multicolumn{1}{c}{21.25$\pm$0.41} & 35.05$\pm$0.56 & \multicolumn{1}{c}{16.06$\pm$0.44} & 27.68$\pm$0.66 & \multicolumn{1}{c}{23.85$\pm$0.40} & 38.52$\pm$0.52 & \multicolumn{1}{c}{28.84$\pm$0.14} & 44.77$\pm$0.17 \\ \hline
        
        \multirow{3}{*}{RNN-Based} 
        & LSTM-\textit{flat} & \multicolumn{1}{c}{27.50$\pm$1.09} & 42.54$\pm$1.22 & \multicolumn{1}{c}{18.07$\pm$0.44} & 30.18$\pm$0.64 & \multicolumn{1}{c}{31.31$\pm$1.33} & 47.18$\pm$1.59 & \multicolumn{1}{c}{32.69$\pm$1.71} & 48.55$\pm$1.18 \\ 
        & LSTM-\textit{hier} & \multicolumn{1}{c}{30.20$\pm$0.47} & 46.39$\pm$0.56 & \multicolumn{1}{c}{22.86$\pm$0.42} & 37.21$\pm$0.56 & \multicolumn{1}{c}{32.90$\pm$0.93} & 49.51$\pm$1.05 & \multicolumn{1}{c}{36.00$\pm$0.50} & 52.94$\pm$0.54 \\
        & RETAIN\cite{choi2016retain} & \multicolumn{1}{c}{31.16$\pm$0.82} & 42.16$\pm$0.99 & \multicolumn{1}{c}{21.13$\pm$0.64} & 30.49$\pm$0.96 & \multicolumn{1}{c}{36.70$\pm$0.86} & 48.54$\pm$0.73 & \multicolumn{1}{c}{43.19$\pm$1.06} & 54.14$\pm$1.20 \\
        & DAG-ERC\cite{shen2021directed} & \multicolumn{1}{c}{29.08$\pm$0.56} & 44.05$\pm$0.70 & \multicolumn{1}{c}{23.74$\pm$0.76} & 35.71$\pm$1.02 & \multicolumn{1}{c}{36.16$\pm$0.46} & 53.80$\pm$0.52 & \multicolumn{1}{c}{31.18$\pm$1.05} & 47.52$\pm$1.23 \\
        
        \hline
        
        \multirow{3}{*}{Transformer} 
        & HiTANet\cite{luo2020hitanet} & \multicolumn{1}{c}{30.75$\pm$0.69} & 44.57$\pm$0.83 & \multicolumn{1}{c}{22.01$\pm$1.04} & 33.62$\pm$1.44 & \multicolumn{1}{c}{33.95$\pm$1.26} & 48.39$\pm$1.26 & \multicolumn{1}{c}{39.17$\pm$1.93} & 53.41$\pm$2.21 \\
        & LSAN\cite{ye2020lsan} & \multicolumn{1}{c}{34.33$\pm$0.58} & 46.14$\pm$0.45 & \multicolumn{1}{c}{26.11$\pm$1.06} & 38.89$\pm$1.01 & \multicolumn{1}{c}{39.28$\pm$0.22} & 52.49$\pm$0.62 & \multicolumn{1}{c}{50.29$\pm$1.24} & 57.90$\pm$1.09 \\
        & DialogXL\cite{shen2021dialogxl} & \multicolumn{1}{c}{36.27$\pm$0.34} & 53.23$\pm$0.40 & \multicolumn{1}{c}{27.12$\pm$0.24} & 42.67$\pm$0.36 & \multicolumn{1}{c}{40.91$\pm$0.14} & 58.06$\pm$0.15 & \multicolumn{1}{c}{48.68$\pm$0.81} & 65.48$\pm$0.66 \\
        & {\model}(Ours) & \multicolumn{1}{c}{\textbf{42.62$\pm$0.35}} & \textbf{59.77$\pm$0.34} & \multicolumn{1}{c}{\textbf{32.26$\pm$1.25}} & \textbf{48.77$\pm$1.43} & \multicolumn{1}{c}{\textbf{44.86$\pm$0.54}} & \textbf{61.93$\pm$0.52} & \multicolumn{1}{c}{\textbf{56.99$\pm$0.53}} & \textbf{72.60$\pm$0.43} \\ 
        \bottomrule
    \end{tabular}}}
    \caption{Performance~($\%$) comparison of {\model} with baseline methods over the overall and three departments datasets. The best result in each column is highlighted in boldface. The  performance gain of our method over all baselines is statistically significant with p \textless~0.05 under t-test. }
    \label{tab:results}
\end{table*}

\subsection{Experimental Setup}
\paragraph{Dataset}
In our experiments, we divide the data into train/development/test dialogue sets as shown in Table~\ref{tab:dataset}. 
The average number of medications in each dialogue is approximately the same, as well as the the average length of utterances and dialogues, meaning the distribution of the data is relatively consistent among three sets.
\paragraph{Implementation Details}
The pretained model we use is Chinese RoBERTa-base model.
The learning rate and the batch size are set as $2 \times 10^{-5}$ and $8$, respectively. 
Adam optimizer is utilized to optimize the model. All methods are implemented and trained using Pytorch on GeForce RTX 3090 GPUs.
The results are the mean of five trainings.

\paragraph{Baselines}
Since there is no standard baselines for this task,
we implement several methods of related tasks, including statistics-based (i.e., \textbf{TF-IDF~\cite{salton1988term}}), RNN-based (i.e., \textbf{LSTM-\textit{flat}}, \textbf{LSTM-\textit{hier}}, \textbf{RETAIN}~\cite{choi2016retain} and \textbf{DAG-ERC~\cite{shen2021directed}}), and transformer-based methods (i.e., \textbf{HiTANet}~\cite{luo2020hitanet}, \textbf{LSAN}~\cite{ye2020lsan}) and \textbf{DialogXL~\cite{shen2021dialogxl}}. 
The RETAIN, HiTANet and LSAN are strong baselines for EHR-based medication recommendation or risk prediction. DAG-ERC and DialogXL are the SOTA methods at Emotion Recognition in Conversation (\textbf{ERC}).
\textit{Among them, LSTM-\textit{hier} takes the dialogue structure into consideration, and 
LSAN and DialogXL are modified to incorporate disease knowledge}. 
Refer to Appendix \ref{sec:baselinesdetail} for more details.

\paragraph{Evaluation Metrics}
We adopt two commonly used metrics, namely \textbf{Jaccard} and \textbf{F1} scores, to evaluate the model performance.

\subsection{Main Results}
Table~\ref{tab:results} shows performances of all methods under the metric of $\mathrm{Jaccard}$ and $\mathrm{F1}$ on four datasets.
The results clearly indicate that {\model} has achieved the best performances among all baselines.
Particularly, {\model} improves $6.35\%$, $5.14\%$, $3.95\%$, and $8.31\%$ compared with the second best method~(i.e., \textbf{DialogXL}) at $\mathrm{Jaccard}$, respectively.
Further, RETAIN and LSTM-{\textit{hier}} outperform LSTM-{\textit{flat}}, demonstrating the dialogue structure is important for the dialogue understanding.
And LSAN, DialogXL outperforms HiTANet, indicating that disease knowledge is also essential for the dialogue modeling.
Our well-designed model {\model} considers both of the above and achieves the best performance.
In addition, it is worth noting that the performance varies over three departments, which may attribute to the considerable difference of medication and disease frequencies between different departments.
\begin{figure}[t]
    \centering
    \subfigure[Jaccard on four datasets]{
    \includegraphics[scale=0.25]{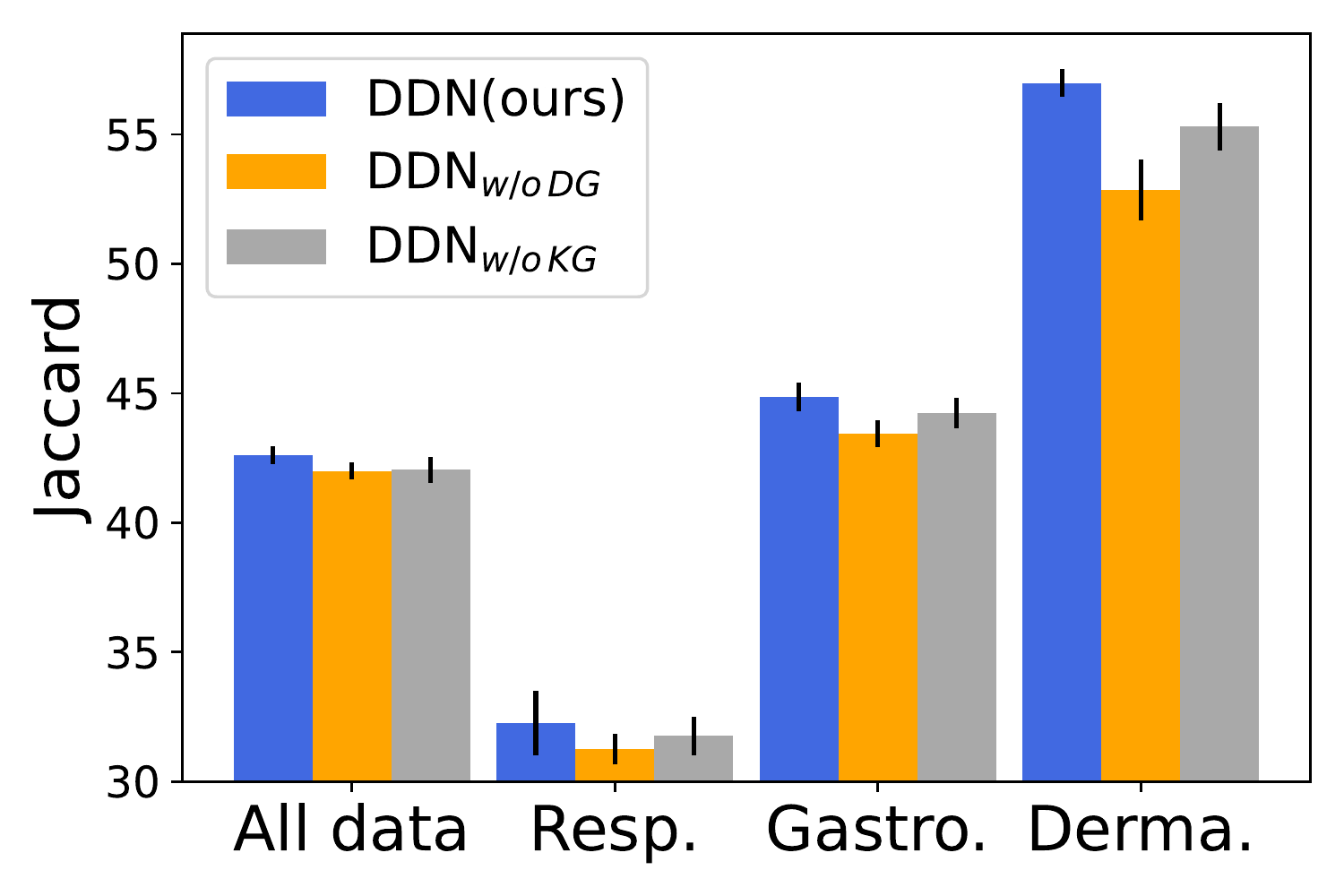}
    }%
    \subfigure[F1 on four datasets]{
    \includegraphics[scale=0.25]{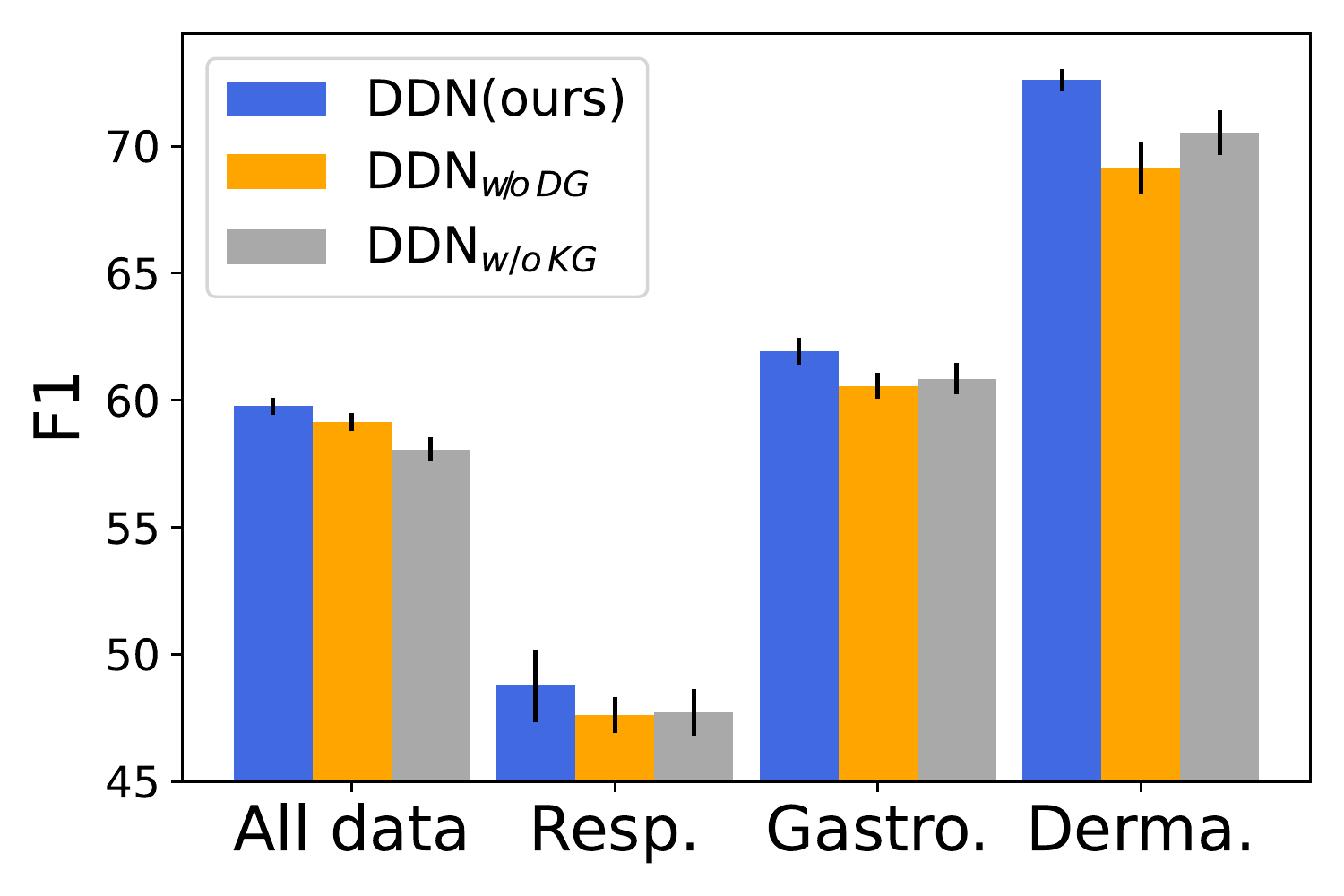}
    }
    \caption{Performance comparison of {\model} and its variants.}
    \label{fig:ablation}
\end{figure}%

\subsection{Ablation Study}
Figure~\ref{fig:ablation} summarizes the contributions of QA Dialogue Graph and disease knowledge of our model.
We notice that by removing the QA Dialogue Graph, the variant {\model}$_{w\!/\!o \, DG}$ shows considerable performance decrease at both Jaccard and F1 compared with {\model}, especially on three departments datasets. It demonstrates that dialogue graph structure is important for the medical information extraction in dialogue-based medication recommendation task.
Similarly, by removing the Knowledge Graph module,  {\model}$_{w\!/\!o \, KG}$ also shows similar performance decrease trends, indicating that disease knowledge can improve the medication recommendation performance. This is reasonable and accords with the actual medication consultation situations.

\subsection{Task Feasibility Analysis}
\begin{figure}[!t]
    \centering
    \includegraphics[scale=0.35]{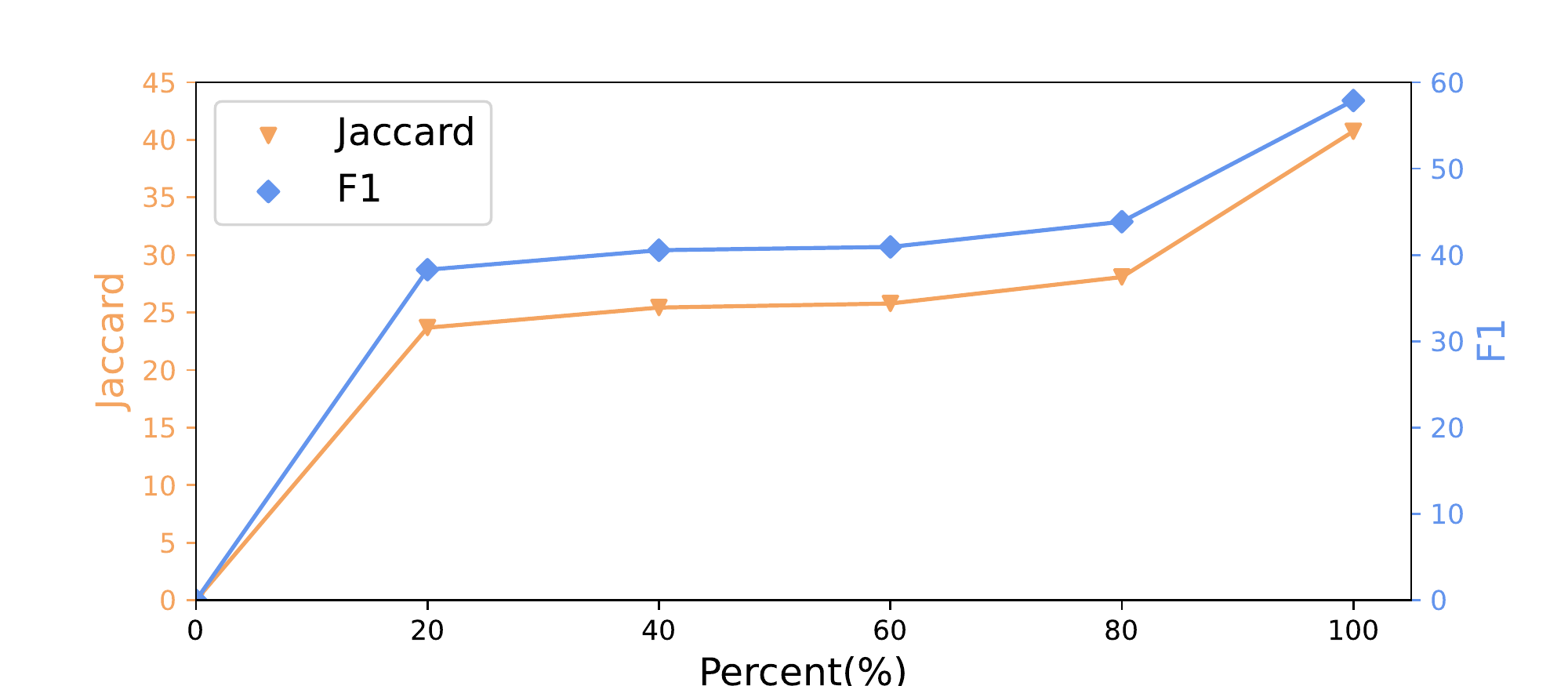}
    \caption{Average Jaccard scores on different percents($\%$) of dialogue discourse. In this setting, we choose dialogues with more than four turns in test set.}
    \label{fig:dialogue_discourse}
\end{figure}%
To prove the feasibility of dialogue-based medication recommendation, we provide incomplete discourses to {\model} during the inference process to explore whether the dialogue can provide necessary medical information.
Figure~\ref{fig:dialogue_discourse} shows the model performances under different portions of discourses.
We can see that with the increasing of dialogue discourse percentage, the performance gets better, especially within the first $20\%$ and the last $20\%$. 
This may be because that the first and last parts of dialogue contain much patient complaints and symptoms that are closely related to the medications.
The results demonstrate that recommending medication based on medical dialogues is feasible.%
\begin{table}[!b]
    \small
    \centering
    \resizebox{0.8\columnwidth}{!}{
        \begin{tabular}{lll}    
        \toprule    
          No. & Type of error & \# Cases  \\
          \midrule 
          \#1 & $\mathit{P} \subseteq \emptyset$ & 65(7.20\%)  \\
          \#2 &  $\mathit{P} \subset \mathit{T}~\&~\mathit{P} \not\subseteq \emptyset$ & 58(6.42\%)  \\ 
           \#3 & $\mathit{T} \subset \mathit{P} $  & 182(20.16\%) \\ 
          \#4 & $\mathit{T} \not\subseteq \mathit{P}~\&~\mathit{P} \not\subseteq \mathit{T}~\&~\mathit{P} \cap \mathit{T} \not\subseteq \emptyset$  & 299(33.11\%) \\
         \#5 &  $\mathit{T} \not\subseteq \mathit{P}~\&~\mathit{P} \not\subseteq \mathit{T}~\&~\mathit{P} \cap \mathit{T} \subseteq \emptyset$ & 299(33.11\%) \\
          \hline
          Total & - & 903 \\
        \bottomrule   
        \end{tabular} 
    }
    \caption{The statistics of errors on test set. $\mathit{P}$ and $\mathit{T}$ are the predicted and golden label set, respectively.}
    \label{tab:error_analysis}
\end{table}

\subsection{Error Analysis}
Although we have elaborately designed a model for the task, the results are not so well satisfactory. So we make detailed analysis of the error cases in the test set.
Table~\ref{tab:error_analysis} summarizes the statistics of our defined five type of errors.
We can see that (1) $86.38\%$ of the cases~(\#3, \#4, \#5) predict wrong medications, which is mainly caused by {\model} failing to distinguish the medications with similar effect. (2) $7.20\%$ of the cases predict none labels, which can be attributed to that these dialogues provide a little disease-related information.

\subsection{Case Study}
\begin{figure}[!tbp]
    \centering
    \includegraphics[scale=0.6]{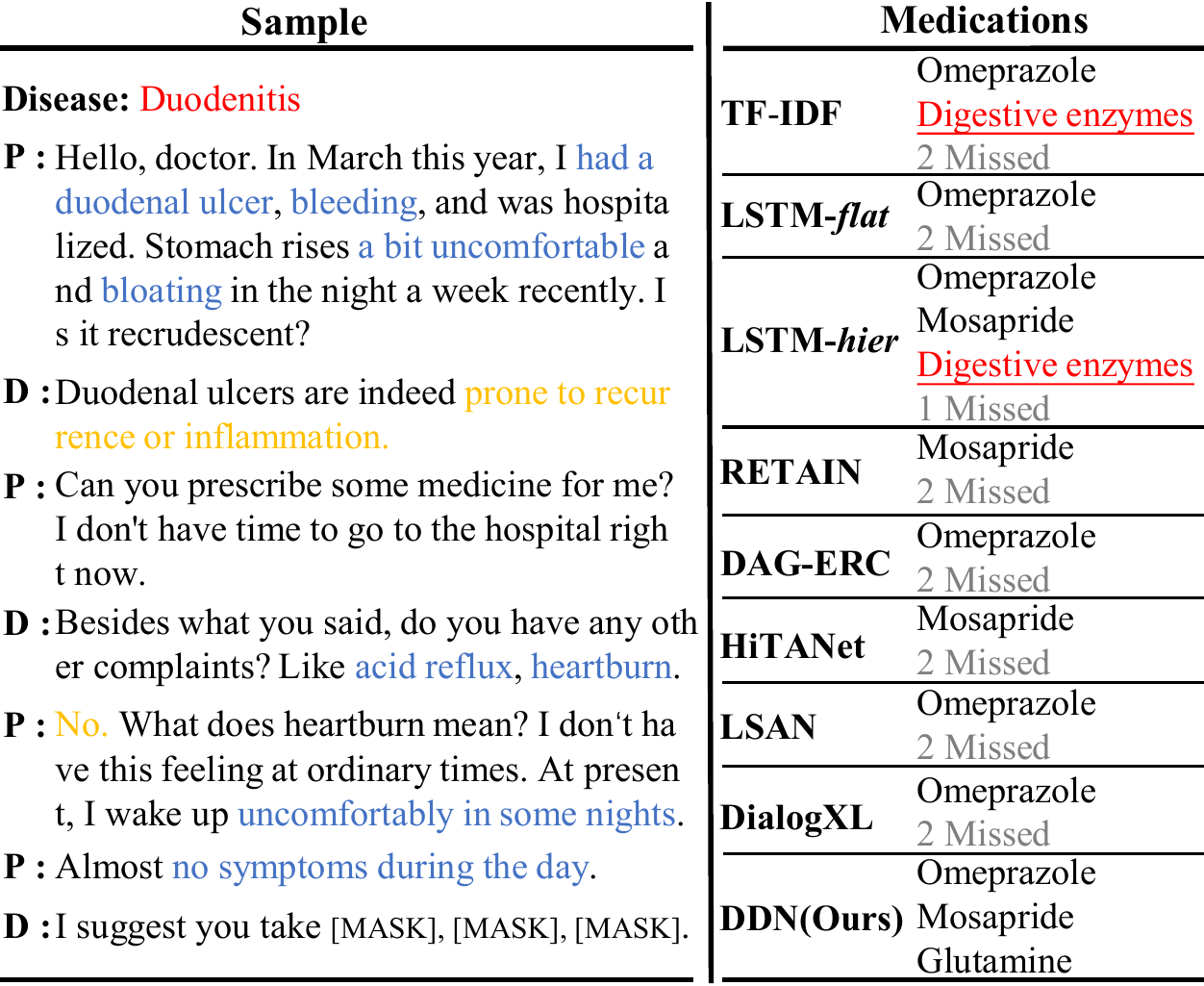}
    \caption{The sample is extracted from the {\dataset} test set. Golden labels of this case are Omeprazole, Mosapride and Glutamine. The "Missed" means the medication is in golden labels but not be predicted, and the underlined drugs in red represent the predicted medications that are not in ground truth.}
    \label{fig:case_study}
\end{figure}

We further provide a case study to illustrate the superiority of {\model}.
Figure~\ref{fig:case_study} shows the medical dialogue and the medications recommended by all baselines and our method.
The baselines either miss some medications, e.g., LSTM-\textit{flat}, RETAIN, HiTANet, LSAN, or give the wrong drugs, e.g., TF-IDF, LSTM-\textit{hier}. 
{\model} takes full account of \textit{Duodenitis}-related information from the dialogue (e.g., the symptoms in chief complaint and past medical history) and the external knowledge graph.
It recommends \textit{Omeprazole} (inhibiting gastric acid secretion) and \textit{Mosapride} (promoting gastric dynamics), as well as \textit{Glutamine} which is omitted by all baselines.
\section{Conclusions}
\label{sec:conclusion}
In this paper, we studied a new task, namely dialogue-based medication recommendation. 
First, we presented the first high-quality medical dialogue dataset {\dataset} for this task.
And then we implemented several baselines, as well as designed a dialogue structure and external disease knowledge aware model. 
Experimental results show that medication recommendation quality can be enhanced with the help of dialogue structure and external disease knowledge.

\section*{Ethical considerations}
\label{sec:ethical}
Data in {\dataset} is publicly collected from Chunyuyisheng, and personal information (e.g., usernames) is preprocessed. The annotating process is as described in Section \ref{sec:corpus}. Furthermore, to ensure the quality of dataset, we paid the annotators $1$ yuan ($\$0.16$ USD) per label. 
The applications of machine learning in medical treatment would inevitably raise ethical issues. 
But the research on AI medicine should not be stopped by this, since the purpose of such research is how to make machines better serve human beings. 
We have seen many advanced achievements~\cite{lin2021graph,10.1145/3404835.3462921,zhang2020mie,liu2020meddg,lin2019enhancing,xu2019end,wei2018task} in this field. 
For this study, the ethical issue is that there may cause bad cases in practical application. However, individual errors could be reduced by making doctors responsible for decisions while machines are used as assistants.

\section*{Acknowledgements}

This research was partially supported by National Key R\&D Program of China under grant No. 2018AAA0102102, National Natural Science Foundation of China under grants No. 62176231 and 62106218, Zhejiang public welfare technology research project under grant No. LGF20F020013.
The authors would  thank Ruochen Yan for her help in data processing.

\bibliography{anthology,custom}
\bibliographystyle{acl_natbib}

\clearpage
\appendix
\section{Corpus}
\label{sec:corpus_append}

\subsection{Details of corpus construction}
\label{sec:detail_process}
First of all, diseases and related medications were identified in a dialogue. Secondly, we selected and annotated those dialogues containing drugs in our medication list. To speed up tagging process, we built an annotation tool based on this task. For each raw medical dialogue, the annotators need to annotate the disease of patients and medications recommended by doctors. We believe that the context after the doctor recommending the drug is not meaningful for drug inference. 
Due to the emergence of new medications in the labeling process and existence of ambiguity on recommendation, two additional annotation processes were carried out. Next we will focus on the 
processing of diseases and medications.

\paragraph{Disease Processing.} With the guidance of a doctor, we select 16 diseases from 3 departments (i.e., respiratory, gastroenterology and dermatology) with following reasons: (1) they are common diseases and research on them have more practical value. (2) they could be consulted online and there are abundant medication consultations. As described by Section \textbf{Corpus Description}, we normalize the diseases to improve the quality of {\dataset}, e.g., \textbf{chronic gastritis} and \textbf{acute gastritis} are mapped to \textbf{gastritis}. The dialogues without explicit disease information or diseases in our scope were marked as \textbf{None or Others}. We mark one disease according to the chief complaint of patients who have more than one disease, because patients have only one complaint in most diagnostic scenarios.

\paragraph{Medication Processing.} As for medications, the ones we choose are commonly prescribed by doctors. Considering the differences between traditional Chinese medicines and Western medicine, both are included to achieve complementary advantages. Since there are many generic names, trade names and colloquial expressions for the same drug in conversations, it is significant to normalize the drug to a single label. 
For example, \textbf{Omeprazole enteric-coated tablet} and \textbf{Omeprazole enteric-coated capsule} could be mapped to \textbf{Omeprazole}. For compound medicines, we combine drugs that have the same ingredients into one, e.g., Tylenol represents all medicines that contain acetaminophen, pseudoephedrine hydrochloride, dextromethorphan hydrobromide and chlorpheniramine maleate. Due to space constraints, more normalization of diseases and medications could be found in  our repository\footnote{https://github.com/Hhhhhhhzf}.

\section{Experiments} 

\subsection{Baselines}
\label{sec:baselinesdetail}
\begin{itemize}
\item \textbf{TF-IDF}. This is a traditional bag-of-word model for text classification. We view each dialogue as text and the corresponding medication as label, and train a classification model based on TF-IDF features of words.
\item \textbf{LSTM-\textit{flat}}. This is a LSTM-based method. It concatenates all the sentences in a dialogue as a long sentence and feeds the long sentence into the BiLSTM to get the dialogue embedding for medication prediction.
\item \textbf{LSTM-\textit{hier}}. This is also a LSTM-based method. Different from LSTM-\textit{flat}, it uses a hierarchical BiLSTM where each word in an utterance are fed into BiLSTM to get the utterance embedding and then the utterances are fed into another BiLSTM to get the final dialogue embedding. It captures both word-level and utterance-level dependencies. 
\item \textbf{RETAIN}. This is a RNN-based EHR medication recommendation method using on a two-level neural attention network that detects influential past visits. In the current scenario, it is used to model the dialogues.
\item \textbf{DAG-ERC}. This method designed a directed acyclic neural network to model the information flow between long-distance conversation background and nearby context. Following the implementation in~\cite{shen2021directed}, the features of utterances extracted from fine-tuning RoBERTa are inputted in model while the model structure is RNN based, so DAG-ERC is regraded as a RNN-based model.
\item \textbf{HiTANet}. This is a Transformer-based risk prediction approach on EHR, which model time information in local and global stages. We transform this method to model the hidden temporal information in medical dialogues.
\item \textbf{LSAN}. This is also a Transformer-based risk prediction approach, to model the hierarchical structure of EHR data. We modified this method to model the hierarchical structure in medical dialogues and add disease module of {\model} to encoder the external knowledge.
\item \textbf{DialogXL}. This method improves XLNet with enhanced memory and dialog-aware self-attention. We modify the softmax layer to sigmoid layer in this model to fit the multi-label task in medication recommendation and add the disease module of {\model}.

\item \textbf{{\model}}. This is our proposed model. It utilizes the dialogue structure and external disease knowledge to enhance the dialogue-based medication recommendation performance.
\end{itemize}

\subsection{Evaluation Metrics}
\label{sec:metrics}
\begin{gather}
\text{Jaccard}=\frac{1}{|D|}\sum_{k=1}^{|D|}\frac{|Y^{(k)}\cap \hat{Y} ^{(k)}|}{|Y^{(k)}\cup \hat{Y} ^{(k)}|} \\
\text{F1}=\frac{1}{|D|}\sum_{k=1}^{|D|}\frac{2 \cdot \text{P}^{(k)}\cdot \text{R}^{(k)}}{\text{P}^{(k)}+\text{R}^{(k)}}
\end{gather}
where $|D|$ is the number of dialogues in the test set. $Y^{(k)}$ represents the ground truth medication set of the $k$th dialogue, and $\hat{Y} ^{(k)}$ represents the predicted medication set of the $k$th dialogue by the model. 
$\text{P}^{(k)}$, $\text{R}^{(k)}$ represents the Precision and Recall of the $k$th dialogue, respectively.

\subsection{Additional Experiment on DDI}
\label{sec:ddi}
Medication combination recommendation would trigger the Drug-Drug Interaction (\textbf{DDI}) inevitably, which might lead to adverse outcomes. To this end, we explore the DDI in {\dataset}. 
And we follow the previous work \cite{shang2019gamenet} to give the DDI rate definition~(smaller value means better).
\begin{equation}
    \small
    DDI Rate = \frac{\sum_{k}^{N} \sum_{i, j} | \{( c_i, c_j ) \in \hat{Y}^{(k)}|( c_i, c_j ) \in \mathcal{E}_d \}|}{\sum_{k}^{N} \sum_{i, j} 1}
\end{equation}
where the set will count each medication pair $(c_i, c_j)$ in recommendation
set $\hat{Y}$ if the pair belongs to edge set $\mathcal{E}_d$ of the
DDI graph. Here $N$ is the size of test dataset. In addition, DDI relationships among medications in {\dataset} are collected from YAOZH \footnote{\url{https://db.yaozh.com/interaction}}, a medical data retrieval system. 

The evaluation results are shown in Table \ref{tab:ddi_results}. We could find that ground truth DDI rate is very small~(compared to the 8.08\% in MIMIC-III~\cite{yang2021safedrug}), which may lead to the low rate on models. In view of this situation, we think it is no need for additional efforts to control the DDI rate at the current stage. 
Considering for the future research, we open source our DDI relationship graph in our repository.
\begin{table}[h]
    \centering
    \small
    \resizebox{0.99\linewidth}{!}{
    \begin{tabular}{ccccc}
    \toprule
         & \multicolumn{4}{c}{\textbf{DDI Rate}} \\
         \midrule
        Model & All Data & Respiratory & Gastroenterology & Dermatology \\
        \midrule
        G.T. & 1.12 & 0.78 & 2.06 & 0.74 \\
        TF-IDF & 1.10 & 0.46 & 2.01 & 0.51 \\
        LSTM-\textit{flat} & 0.58 & 1.36 & 0.93 & 0.00 \\
        LSTM-\textit{hier} & 1.02 & 0.11 & 0.91 & 0.65 \\
        RETAIN & 1.92 & 1.12 & 1.89 & 0.00 \\
        DAG-ERC & 0.81 & 1.01 & 1.53 & 0.48 \\
        HiTANet & 0.45 & 1.49 & 1.09 & 0.50 \\
        LSAN & 1.57 & 0.00 & 1.62 & 0.48 \\
        DialogXL & 1.34 & 1.09 & 1.59 & 0.40 \\
        DDN & 1.90 & 0.20 & 1.54 & 0.47 \\
    \bottomrule
    \end{tabular}
    }
    \caption{DDI Rate~(\%) comparison on {\dataset}. G.T. represents the Ground Truth.}
    \label{tab:ddi_results}
\end{table}

\section{Task}
\label{sec:task}

\subsection{Medical Utility}
Medical treatment includes a number of steps: registration, examination, image reading, report interpretation, diagnosis, prescription and so on. AI medicine could help optimize resource allocation and improve efficiency in all aspects of health care. To this end, there are two kinds of computer aided diagnosis system, image diagnosis and text diagnosis. Due to the higher threshold of diagnosis, current researches are more inclined to image analysis, and there is still a lot of room for development in text diagnosis. Conversations in outpatient clinics are not reserved and involved many severe data privacy implications, leading to dialogue-based drug recommendation mainly oriented to telemedicine. The medical dialogue system, as a assistant of doctors, could give auxiliary medication suggestions based on the contexts when doctors and patients are communicating with each other.

\section{Statistics}
\label{sec:statistics}

\subsection{Ratio of consulting for medications}
\label{sec:ratio_medication}
The ratio of the patients to consult for medications is calculated with regular expressions. In the first place, 10,0000 different medical conversations from our dialogue corpus based on random sampling are fetched. For every dialogue, we apply the regular expression (e.g., "[Ww]hat (medication$\mid$drug$\mid$medicine) should I (take$\mid$eat)") on the utterances spoken by the patient and assume that it is a case of consulting for drugs if the regular expression matches. The regular expressions are collected based on our observation and understanding of data. More regular expressions could be found in our repository.

\subsection{Complete Corpus Statistics}
\label{sec:complete_corpus_statis}
\begin{figure}[!h]
    \centering
    \includegraphics[scale=0.45]{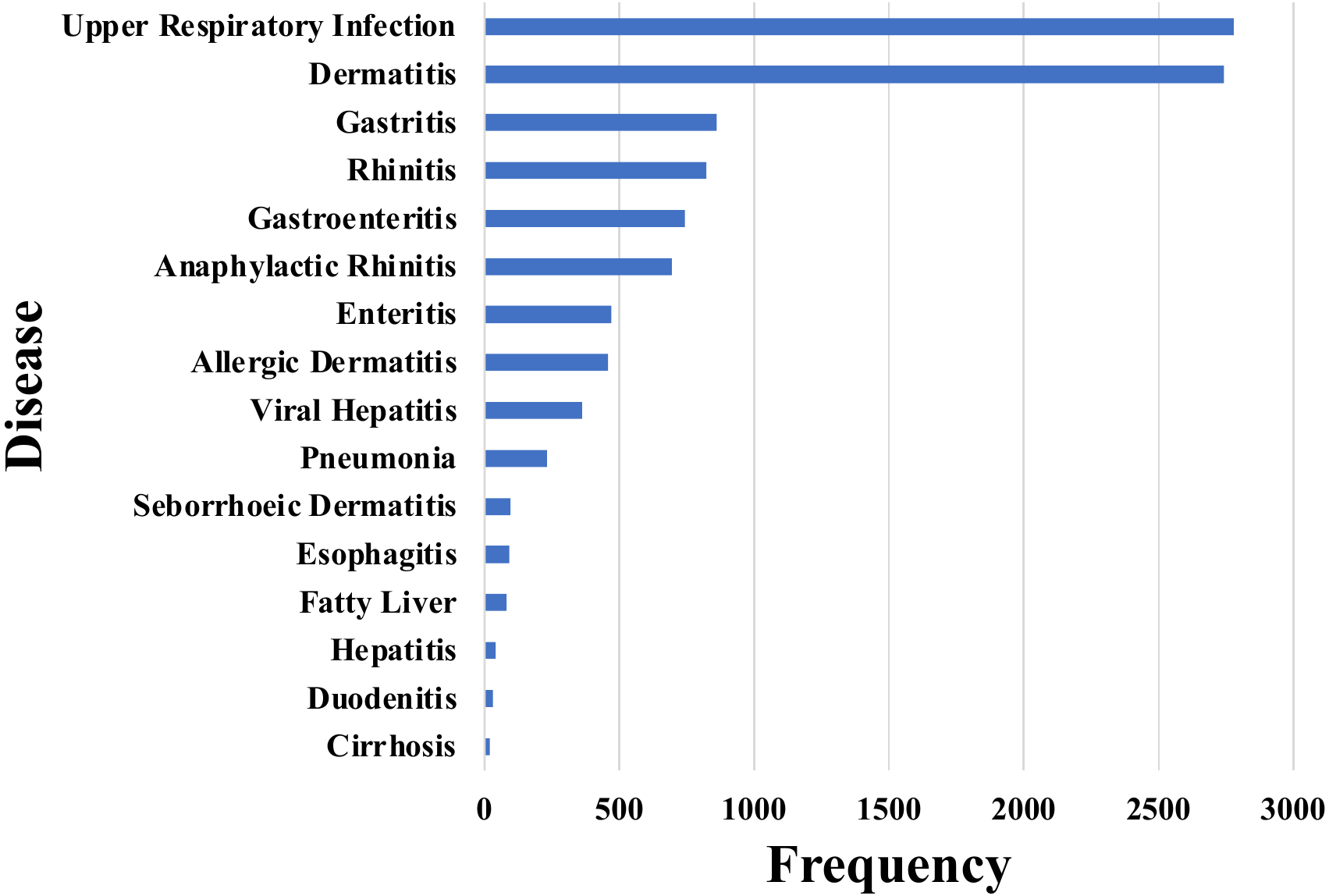}
    \caption{The frequency of all  diseases.}
    \label{fig:all_diseases}
\end{figure}

\begin{figure}[!htbp]
    \centering
    \includegraphics[scale=0.45]{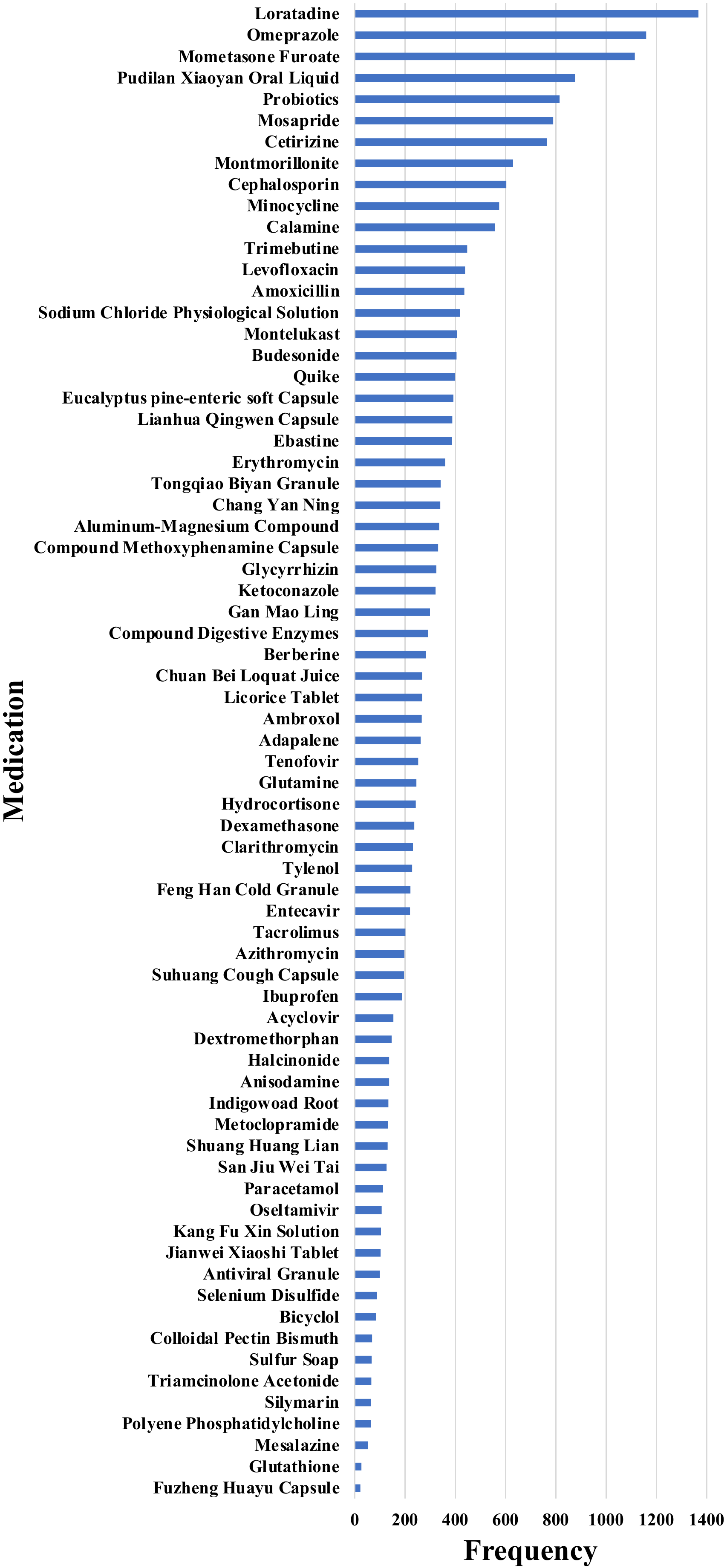}
    \caption{The frequency of all medications. The names are translated from Chinese.}
    \label{fig:all_medications}
\end{figure}

The frequency of all diseases and medications is shown in Figure \ref{fig:all_diseases} \& \ref{fig:all_medications}.

\end{document}